\documentclass[final]{cvpr}

\usepackage{times}
\usepackage{epsfig}
\usepackage{graphicx}
\usepackage{amsmath}
\usepackage{amssymb}

\usepackage{multirow}
\usepackage{subfigure}

\usepackage[pagebackref=true,breaklinks=true,colorlinks,bookmarks=false]{
hyperref}

\def\cvprPaperID{33} % *** Enter the CVPR Paper ID here
\def\confYear{CVPR 2021}

\begin{document}
\pagestyle{empty}

%%%%%%%%% TITLE
\title{Filtering Empty Camera Trap Images in Embedded Systems}

\author{Fagner Cunha, Eulanda M. dos Santos, Raimundo Barreto, Juan G. Colonna\\
Federal University of Amazonas\\
Manaus, Amazonas, Brazil\\
{\tt\small \{fagner.cunha, emsantos, rbarreto, juancolonna\}@icomp.ufam.edu.br}

}

\maketitle

%%%%%%%%% ABSTRACT
\begin{abstract}
Monitoring wildlife through camera traps produces a massive amount of images, 
whose a significant portion does not contain animals, being later discarded.
Embedding deep learning models to identify animals and filter these images 
directly in those devices brings advantages such as savings in the storage 
and transmission of data, usually resource-constrained in this type of 
equipment. In this work, we present a comparative study on animal recognition 
models to analyze the trade-off between precision and inference latency on edge 
devices. To accomplish this objective, we investigate classifiers and object 
detectors of various input resolutions and optimize them using quantization and 
reducing the number of model filters. The confidence threshold of each 
model was adjusted to obtain 96\% recall for the nonempty class, since instances 
from the empty class are expected to be discarded. The experiments show that, 
when using the same set of images for training, detectors achieve superior 
performance, eliminating at least 10\% more empty images than classifiers 
with comparable latencies. Considering the high cost of generating labels for 
the detection problem, when there is a massive number of images labeled for 
classification (about one million instances, ten times more than those available 
for detection), classifiers are able to reach results comparable to detectors 
but with half latency.~\footnote{Code and models are publicy available at
\url{https://github.com/alcunha/filtering-empty-camera-trap-images}}
\end{abstract}

\thispagestyle{empty}

%%%%%%%%% BODY TEXT
\section{Introduction}

The use of camera traps is a strategy for passive wildlife monitoring that 
involves installing cameras with presence sensors that, when triggered, 
activate a process of recording short sequences of images or videos of animals. 
The objective aimed on this strategy is to generate data showing animal in 
their daily lives, without interfering with the animals' natural behavior 
\cite{norouzzadeh2018automatically}. However, it is very common to have a very 
large number of images collected with no  animals (empty) due to sensor 
accidental triggering, e.g. about 75\% of the images from the Snapshot Serengeti 
dataset do not contain animals~\cite{dryad_5pt92}.

In recent years, several studies have investigated the use of deep learning
models to recognize animals in camera trap 
images~\cite{beery2018recognition, norouzzadeh2018automatically, 
schneider2020three, tabak2019machine, villa2017towards, willi2019identifying}. 
Due to the development of new technologies for this type of monitoring, 
embedding models to identify animals directly in the devices may provide 
advantages. For instance, in projects whose equipment is connected to a network, 
prior filtering avoids the unnecessary transmission of empty images, saving 
network bandwidth and energy~\cite{elias2017s}. Another example are traditional 
approaches, which usually involve going to the capture points to retrieve 
cameras and/or memory cards~\cite{team_protocol}. In this case, discarding empty 
images can save data storage, extending the time a camera can stay in the field 
collecting images without needing maintenance.

Although there are studies comparing the performance of several modern 
deep learning architectures on recognizing animals in camera trap 
images~\cite{norouzzadeh2018automatically, schneider2020three, 
villa2017towards}, there are few studies regarding the trade-off between 
model performance and inference latency directly on edge devices.
In~\cite{tyden2020edge}, Tydén and Olsson evaluate object detection models on a 
Raspberry Pi applied to recognize animals. However, the used image dataset is 
composed of 4,000 instances comprising only 8 species, limiting the scope of 
the conclusions. In a similar strategy, Zualkernan 
\etal~\cite{zualkernan2020towards} evaluate several classifiers on a Raspberry
Pi aiming to identify animals. Nonetheless, they do not detail latency 
assessment. Finally, Schneider \etal~\cite{schneider2020three} suggest that 
detection models would be better than classification models for recognizing 
animals, but requiring higher computational power on the other hand. Thus, this 
work presents a comparative study between classifiers and detectors of different 
complexities running on edge devices to recognize nonempty images.

Our main findings are as follows:
\begin{itemize}
 \item Detection models generally outperform classifiers with 
comparable latency if they are trained on the same camera trap image set.
 \item Classifiers can obtain high performance on empty image identification,
 given the difficulty of generating labels to train 
detectors. The performance, however, depends on the amount of images 
available for training, as well as on specific factors that can vary from 
dataset to dataset.
 \item Models with a higher input resolution perform better. Their inference 
latency can be kept low by reducing the number of model filters at an acceptable 
performance cost for the problem.
\end{itemize}

\section{Related Work}

Several works apply deep learning techniques to recognise animals in 
camera trap images. Some studies classify the species assuming that an animal 
has already been identified on the scene~\cite{villa2017towards}. A second 
approach adds an extra class to the model to identify empty images, sometimes
called negative class or background class~\cite{beery2018recognition, 
schneider2020three, tabak2019machine}. There are also approaches composed of 
two stages. In the first stage, a model recognizes whether there are animals in 
the scene, while in the second stage another model is  responsible for the 
species identification~\cite{norouzzadeh2018automatically, 
willi2019identifying}.

From a different perspective, animal detection-based approaches first focus on
localizing animals in images and further classifying their species. In this 
context, MegaDetector~\cite{beery2019efficient} is a model based on the Faster 
R-CNN~\cite{ren2015faster} object detector trained on a large number of camera 
trap images whose objective is to be a generalist animal detector. MegaDetector
was trained to be capable of localizing animals in images from different 
ecosystems in the world, even species not seen during its training. Its primary 
idea is not species identification but only to find and localize animals. The
species identification task must be performed by a classifier trained 
specifically for each project. However, since not all images used for training 
MegaDetector are publicly available due to licensing restrictions, it is not 
possible to reproduce the same experimental conditions as 
in~\cite{beery2019efficient}. In addition, Faster R-CNN requires high 
processing power and is not suitable for running on edge 
devices~\cite{sandler2018mobilenetv2}.

In line with the idea of reducing complexity, some studies compare the accuracy
of Convolutional Neural Networks of varied complexities in the task of 
extracting information from camera trap images using classification 
models~\cite{norouzzadeh2018automatically, schneider2020three, tyden2020edge, 
villa2017towards}. Norouzzadeh \etal~\cite{norouzzadeh2018automatically} 
trained several architectures to recognize nonempty images. In their 
experiments, the VGG16 architecture reached the best result, precisely 96.8\% of 
accuracy rate. It is noteworthy, however, that ResNet18 achieved an accuracy of 
96.3\%, just 0.5\% below the accuracy attained by VGG16, and equal to or higher 
than other deeper models in the same ResNet family. On the other hand, even 
though ResNet18 is less complex, it was not designed for inference on edge 
devices.

The scenario is different in~\cite{schneider2020three}, where Schneider \etal 
evaluated more efficient and lightweight deep models, including the MobileNetV2 
and NASNetMobile architectures, in animal species classification. These models, 
designed specifically for computationally limited devices, reached accuracy 
2.5\% and 5\% (in absolute values) lower than accuracy attained by DenseNet201 - 
the high-complex baseline. As expected, DenseNet201 achieved the best result: 
accuracy of 95.6\% when tested on images from the same training locations; and 
68.7\% on images from locations not trained on.

Despite investigating more efficient and/or lightweight deep models, previously
mentioned works did not focus on carrying out tasks on computationally limited 
devices. This is precisely the objective aimed on Tydén and Olsson's 
work~\cite{tyden2020edge}. These authors study the trade-off between 
computational performance and accuracy of the SSD~\cite{liu2016ssd} detector and 
its optimized version SSDLite~\cite{sandler2018mobilenetv2} (using InceptionV2 
and MobileNetV2 as backbones) to recognize animals using the Raspberry Pi 
development board. However, the dataset investigated is composed of 
approximately only 4,000 images distributed among 8 classes, which may impose a 
limitation to the scope of the conclusions provided.

Finally, in terms of classifiers, Zualkernan \etal~\cite{zualkernan2020towards}
evaluate InceptionV3, DenseNet121, ResNet18 and MobileNetV2 models on animal 
recognition using a dataset composed of 34,000 images. Despite comparing all 
these models in terms of accuracy, only InceptionV3 was evaluated on a Raspberry 
Pi 4B to verify inference latency. 

Therefore, in this work, several architectures developed specifically for low
computational power devices are analyzed to perform the task of animal 
recognition in camera trap images. Instead of focusing on a specific set of 
species, we analyse in this work the models' ability to recognize animals, 
regardless of their species, similar to the process proposed for 
MegaDetector~\cite{beery2019efficient}. In this case, we test images 
from the same training sites and from new locations, as it is also done in 
\cite{schneider2020three}. However, unlike the latter, which investigated 
only classification models, in this work we compare classification and detection 
models. In addition, we evaluate other optimization approaches, such as 
quantization and reducing the number of model filters.

\section{Materials and Methods}
Two datasets are investigated in this paper: 1) Caltech Camera Traps; and 2)
Snapshot Serengeti (SS). These datasets are described below.

\subsection{Datasets}

\textbf{Caltech Camera Traps~\cite{beery2018recognition}:} It contains 243,100
images taken from 140 capture locations in the Southwestern United States. The 
instances were labeled in 22 categories, and about 66,000 bounding boxes 
localizing animals were also provided. For our experiments, we selected a subset 
of images from the empty and nonempty classes, grouping into the nonempty 
class images from all other categories that have bounding boxes. Then, 
the dataset was partitioned into training and validation set according to the 
locations recommended in~\cite{caltech_lila}. Moreover, a subset of the 
training partition was split, consisting of 20 locations chosen at random, to be 
used to adjust the training hyper-parameters (called here validation dev). Due 
to the fact that some locations concentrate a large number of images of the 
empty class, which may lead the models to be biased in certain backgrounds, we 
decided to limit to 1000 instances per location the number of empty class 
instances in both training and validation dev partitions. 
Table~\ref{table:caltech_instances} summarizes the number of instances obtained.

\begin{table}[htb]
\begin{center}
\begin{tabular}{l|c|c|c}
\hline
\textbf{Class}   & \textbf{Training} & \textbf{Val\_dev} & \textbf{Validation} 
\\ 
\hline \hline
Empty    & 8574            & 2824              & 19892              \\
Nonempty & 32032           & 3877              & 23410              \\ 
\hline
\end{tabular}
\end{center}
\caption{Number of images used from the Caltech dataset. The empty class of
training and validation dev partitions was limited to 1000 instances per 
location. The nonempty class is composed of images with bounding boxes from all 
other categories.}
\label{table:caltech_instances}
\end{table}

\textbf{Snapshot Serengeti~\cite{dryad_5pt92}:} This dataset 
currently contains more than 7 million camera trap images collected over 11 
seasons in the Serengeti National Park, Tanzania. In this work, we use images 
from the first six seasons in order to keep consistency with previous works using the 
Snapshot Serengeti dataset~\cite{norouzzadeh2018automatically, villa2017towards, 
willi2019identifying}. These images were divided into training and validation 
sets according to the locations (SS-Site), as recommended in~\cite{ss_lila}. As 
was done for the Caltech dataset, we have also created  a validation dev split 
for hyper-parameter adjustment. This partition was obtained by randomly 
selecting 23 locations from the training set. In addition to the SS-Site 
configuration, we have also performed a partitioning by time (SS-Time) taking 
into account that the models could be used in images from new seasons. To 
generate SS-Time, the first four seasons were grouped into a training set, while 
the fifth season was used as validation dev, and the sixth season as the 
validation set. Finally, the dataset was adapted to represent a  bi-class 
problem, where instances from the blank category were labeled to the empty class 
and the others were used to compose the nonempty class. We also balanced the 
classes for the training partition. 

In addition, since there are approximately only 78,000 images from the nonempty
class annotated with bounding boxes, we also perform experiments using subsets 
of instances (called small), which are balanced for the training and validation 
dev partitions. The objective here is to perform a fair comparison between 
classifiers and detectors when both are trained using the same number of images. 
Unless otherwise specified, the experiments conducted in this work use these 
subsets with fewer images. Details about all data partitions of the Snapshot 
Serengeti dataset investigated in this work are shown in 
Table~\ref{table:serengeti_instances}.

\begin{table}[htb]
\begin{center}
\setlength{\tabcolsep}{4pt}
\begin{tabular}{ll|c|c|c}
\hline
                                                                            & 
\textbf{Class}   & \multicolumn{1}{l|}{\textbf{Training}} & 
\multicolumn{1}{l|}{\textbf{Val\_dev}} & 
\multicolumn{1}{l}{\textbf{Validation}} 
\\ \hline \hline
\multirow{2}{*}{SS-Site}                                                    & 
Empty    & 524804                               & 278578               
 
                 & 535817                                 \\
                                                                            & 
Nonempty & 523891                               & 84531                
 
                 & 209183                                 \\ \hline
\multirow{2}{*}{\begin{tabular}[c]{@{}l@{}}SS-Site\\ 
(small)\end{tabular}} & 
Empty             & 51281                                & 6041                 
 
                 & 535817                                 \\
                                                                            & 
Nonempty          & 52081                                & 6041                 
 
                 & 209183                                 \\ \hline \hline
\multirow{2}{*}{SS-Time}                                                    & 
Empty             & 516635                               & 588406               
 
                 & 383981                                 \\
                                                                            & 
Nonempty          & 516630                               & 225647               
 
                 & 75328                                  \\ \hline
\multirow{2}{*}{\begin{tabular}[c]{@{}l@{}}SS-Time\\ 
(small)\end{tabular}} & 
Empty             & 43612                                & 18957                
 
                 & 383981                                 \\
                                                                            & 
Nonempty          & 44579                                & 18957                
 
                 & 75328                                  \\ \hline
\end{tabular}
\end{center}
\caption{Number of instances used from the Snapshot Serengeti dataset.}
\label{table:serengeti_instances}
\end{table}

\subsection{Architectures}

The MobileNetV2~\cite{sandler2018mobilenetv2} architecture is a natural choice
to be used as a classifier baseline, since it was designed specifically for 
computationally limited devices. Besides the original MobileNetV2 model (input 
resolution $224 \times 224$), a version with input $320 \times 320$ was also 
used in our experiments due to the fact that animals can appear very far from 
the camera and visually small as a 
consequence~\cite{norouzzadeh2018automatically}. 

Recently, a new family of models called EfficientNet~\cite{tan2019efficientnet} 
was developed,  whose input resolution, depth and number of filters are scaled 
together from the base model. These models obtained superior performance on 
ImageNet~\cite{ILSVRC15} when compared to models with a similar number of 
parameters and FLOPS, reaching the state of the art with the most complex 
version (EfficientNet-L2). In this work, the EfficientNet-B0 ($224 \times 224$) 
and EfficientNet-B3 ($300 \times 300$) versions were chosen, since they have 
input resolution comparable to MobileNetV2's.

For the detectors, we investigate the SSDLite~\cite{sandler2018mobilenetv2} 
with a MobileNetV2 ($320 \times 320$ input) as backbone. Considering that a new 
family of detectors was recently developed using EfficientNets as 
backbones~\cite{tan2020efficientdet}, we also included in our experiments the 
EfficientDet-D0, which works with an input resolution $512 \times 512$, the 
smallest of this group of models.

\subsection{Implementation Details}

\textbf{Image preprocessing for classification:} We chose a standard procedure 
for image preprocessing. Initially, a random rectangular crop of the image is 
applied with aspect ratio and area sampled in $[3/4, 4/3]$ and $[65\%, 100\%]$, 
respectively. Then, each image is scaled to the input size of each architecture 
and a horizontal flip is applied with $50\%$ probability. Next, we apply data 
augmentation using RandAugment~\cite{cubuk2020randaugment} with parameters $N=2$ 
and $M=2$. Finally, the pixel values of the image are scaled in $[-1, 1]$ for 
MobileNetV2 and in $[0, 1]$ for EfficientNets. During validation, the 
preprocessing consists of only image resizing and pixel scaling depending on the 
architecture.

\textbf{Classifiers training procedure:} Each model was initialized with 
ImageNet pre-trained weights and then trained during 10 epochs using the 
Stochastic gradient descent (SGD) on an NVIDIA GeForce GTX 1080 Ti graphics 
card. The initial learning rate was 0.01 for a batch size of 256 and scaled 
linearly according to the batch size effectively used, which varied according
to the model due to the graphics card limited memory, as shown in 
Table~\ref{table:classifiers_hparams}. The learning rate was 
linearly increased from 0 to the initial rate during 30\% of the steps of the 
first epoch and then reduced using the cosine decay scheduling, as suggested by 
He \etal~\cite{he2019bag}.

\begin{table}[htb]
\begin{center}
\begin{tabular}{l|c|l}
\hline
\multicolumn{1}{c|}{\textbf{Architecture}} & 
\textbf{\begin{tabular}[c]{@{}c@{}}Batch size\end{tabular}} & 
\textbf{\begin{tabular}[c]{@{}c@{}}Learning rate\end{tabular}} \\
\hline \hline
MobileNetV2 (224)                         & 128                                 
 
                               & 0.005                                          
 
                       \\
MobileNetV2 (320)                         & 64                                  
 
                               & 0.0025                                         
 
                       \\
EfficientNet-B0                           & 32                                  
 
                               & 0.00125                                        
 
                       \\
EfficientNet-B3                           & 16                                  
 
                               & 0.000625                                       
 
                       \\ \hline
\end{tabular}
\end{center}
\caption{Batch size and initial learning rate used for each classifier. The 
initial learning rate is scaled linearly according to the batch size $b$, 
defined by $0.01 \times b /256$.}
\label{table:classifiers_hparams}
\end{table}

\textbf{Detectors training procedure:} Both detectors were initialized with 
weights pre-trained on COCO~\cite{lin2014microsoft} and then trained on each 
dataset using the Tensorflow Object Detection API~\cite{huang2017speed}. We have
used the standard training procedure for each detector, besides $32$ and $8$ as 
batch size for \mbox{SSDLite+MobileNetV2} and EfficientDet-D0 respectively. The 
learning rate and the number of training epochs were adjusted according to the 
performance measured on the validation dev partition. These values are shown in 
Table~\ref{table:detectors_hparams}. 

\begin{table}[htb]
\begin{center}
\begin{tabular}{l|l|c|c}
\hline
\textbf{Model} & \textbf{Dataset} & 
\textbf{\begin{tabular}[c]{@{}c@{}}Learning\\rate\end{tabular}} & 
\textbf{\begin{tabular}[c]{@{}c@{}}Training\\steps\end{tabular}} \\ 
\hline \hline
\multirow{2}{*}{SSDLite+MNetV2}  & Caltech & 0.008                              
 
                          & 12000                                              
 
           \\
                                 & SS      & 0.004                              
 
                          & 36000                                              
 
           \\ \hline
\multirow{2}{*}{EfficientDet-D0} & Caltech & 0.008                              
 
                          & 50000                                               
           \\
                                 & SS      & 0.001                              
 
                          & 150000                                              
           \\ \hline
\end{tabular}
\end{center}
\caption{Training hyperparameters for detectors. The same hyperparameters were 
used for both time and site partitioning of Snapshot Serengeti (SS).}
\label{table:detectors_hparams}
\end{table}

\subsection{Evaluation Procedure}

In order to compare detectors and classifiers, predictions related to the 
bounding box coordinates were ignored. Thus, only the detection confidence value 
that represents the class with the highest score was used as the detected label.

Taking into account that the confidence threshold can be adjusted to avoid
nonempty images discarding, we decided to use the precision-recall curve as 
a graphical tool to compare the general performance of the models considering 
all possible thresholds. To compare the models effectiveness when discarding 
empty images, i.e., the true negative rate (TNR), the confidence threshold
of each model was adjusted so that the recall for nonempty images was set to 
96\% -- a value reached by models investigated 
in~\cite{norouzzadeh2018automatically}. 

\begin{figure*}
\begin{center}
\subfigure[Caltech]{\includegraphics[width=0.33\textwidth]{
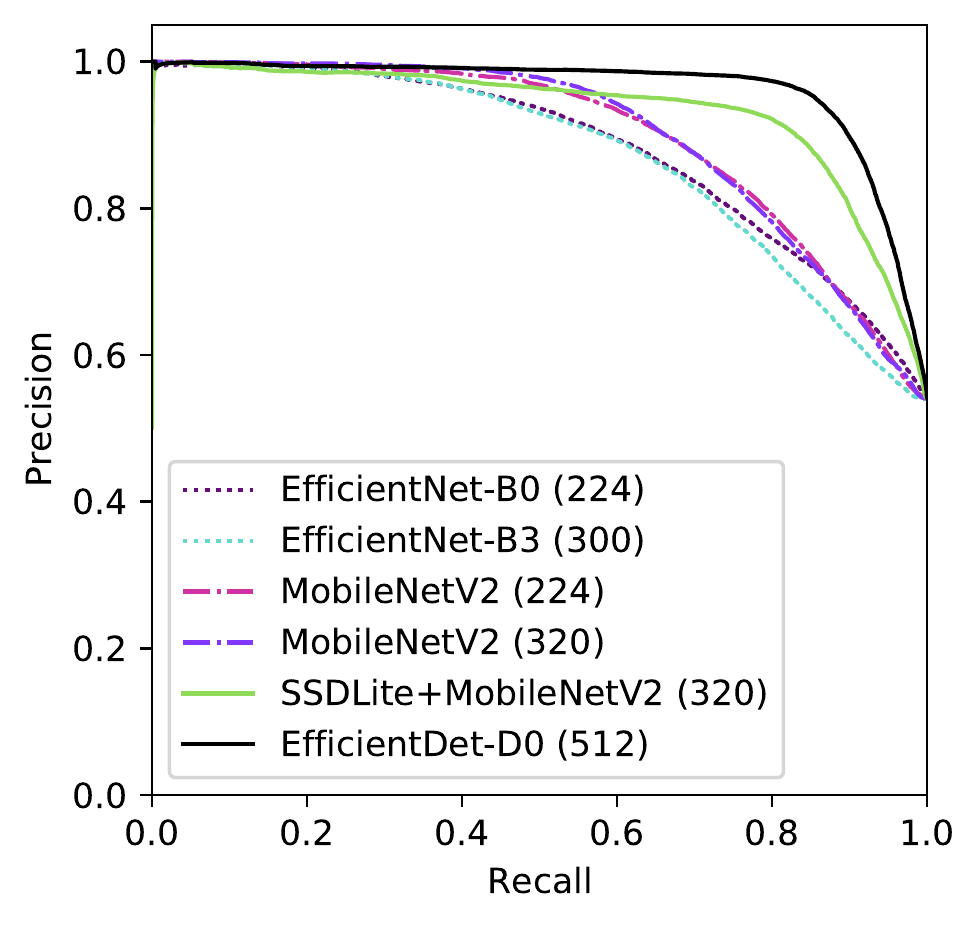}}
\subfigure[SS-Site]{\includegraphics[width=0.33\textwidth]{
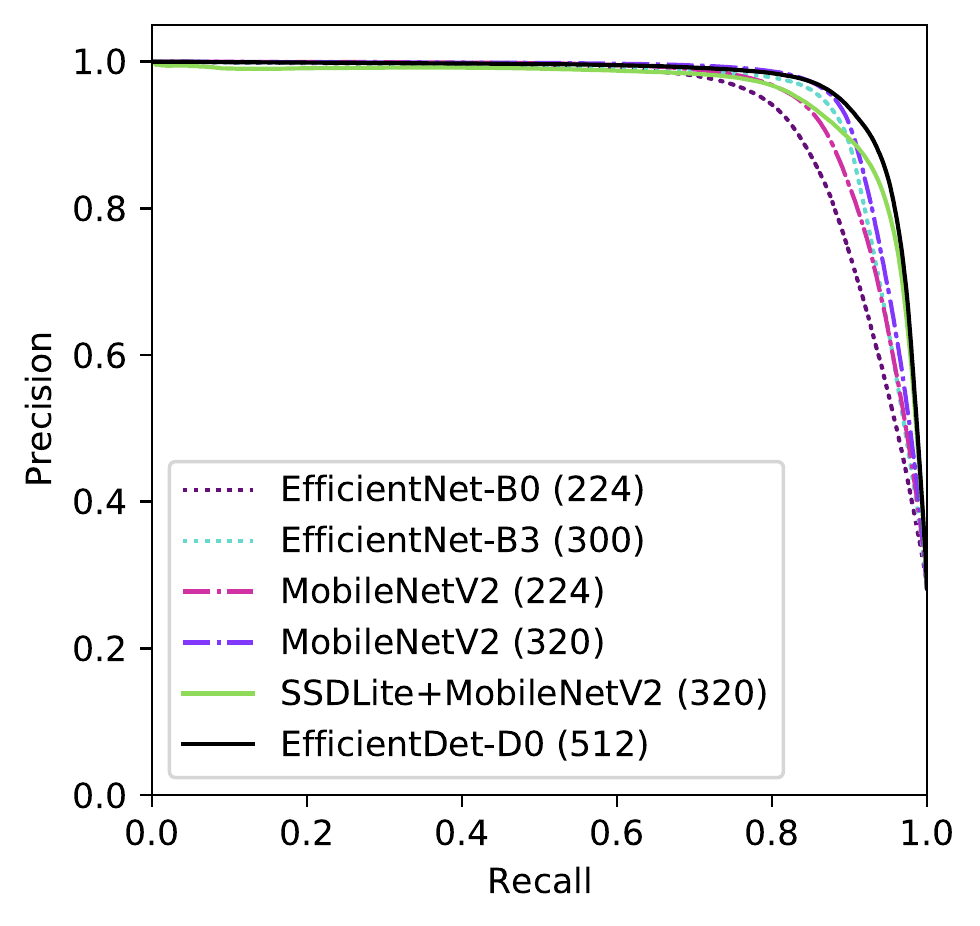}}
\subfigure[SS-Time]{\includegraphics[width=0.33\textwidth]{
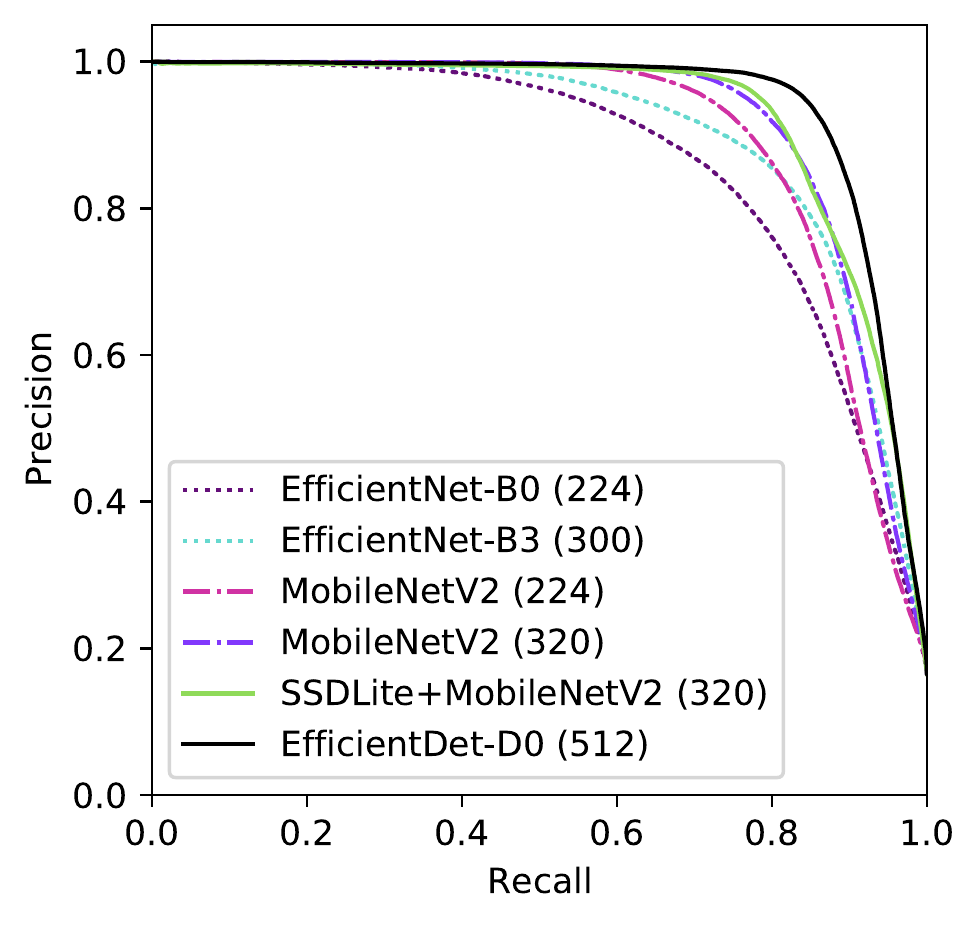}}
\end{center}
   \caption{Precision-recall curve for the nonempty class. Detectors reached
   higher performances, with wider advantage on the Caltech dataset. Best viewed 
in color.}
\label{fig:pr_curve}
\end{figure*}

\begin{table*}[htb]
\begin{center}
\setlength{\tabcolsep}{3pt}
\begin{tabular}{l|c|ccc|ccc|ccc}
\hline
\textbf{}                            &                & 
\multicolumn{3}{c|}{\textbf{Caltech}}                  & 
\multicolumn{3}{c|}{\textbf{SS-Site}}                  & 
\multicolumn{3}{c}{\textbf{SS-Time}}                  \\
\multicolumn{1}{c|}{\textbf{Model}} & \textbf{CPU}   & \textbf{Precision} & 
\textbf{TNR}     & \textbf{Thresh.} & \textbf{Precision} & \textbf{TNR}     & 
\textbf{Thresh.} & \textbf{Precision} & \textbf{TNR}   & \textbf{Thresh.}\\
\hline \hline
Efficientnet-B0                      & 801ms          & \textbf{60.26\%}  & 
\textbf{25.50\%} & 0.355           & 50.32\%           & 63.00\%          & 
0.153           & 33.08\%           & 61.90\%          & 0.159           \\
Efficientnet-B3                      & 3205ms         & 56.42\%           & 
12.75\%          & 0.305           & 57.21\%           & 71.97\%          & 
0.155           & \textbf{39.27\%}  & \textbf{70.87\%} & 0.170           \\
MobileNetV2-224                      & \textbf{324ms} & 58.60\%           & 
20.18\%          & 0.228           & 57.72\%           & 72.55\%          & 
0.188           & 30.51\%           & 57.11\%          & 0.126           \\
MobileNetV2-320                      & 638ms          & 58.58\%           & 
20.13\%          & 0.239           & \textbf{62.84\%}  & \textbf{77.84\%} & 
0.191           & 35.74\%           & 66.13\%          & 0.147           \\ 
\hline
SSDLite+MNetV2                       & \textbf{838ms} & 67.03\%           & 
44.42\%          & 0.166           & 75.32\%           & 87.72\%          & 
0.167           & 47.14\%           & 78.89\%          & 0.147           \\
Efficientdet-D0                      & 4686ms         & \textbf{73.31\%}  & 
\textbf{58.86\%} & 0.148           & \textbf{79.14\%}  & \textbf{90.12\%} & 
0.150           & \textbf{47.35\%}  & \textbf{79.06\%} & 0.143           \\ 
\hline
\end{tabular}
\end{center}
\caption{Comparison of precision for the nonempty class and the true negative 
rate (TNR) where the confidence threshold of each model was adjusted 
to achieve a recall of 96\% on the nonempty class. The reported CPU latency 
corresponds to the Caltech dataset, but it is similar to the others, being 
calculated from the average of 50 runs. The true negative rate indicates the 
percentage of images without animals that would no longer be stored 
unnecessarily.}
\label{table:precision_tvn}
\end{table*}

\begin{figure*}
\begin{center}
\subfigure[EffNet-B0: 0.70, EffNet-B3: 0.67, MNetV2-224: 0.73, MNetV2-320: 
0.60, SSDLite: 0.11, EffDet-D0: 0.13]{\includegraphics[width=0.32\textwidth]{
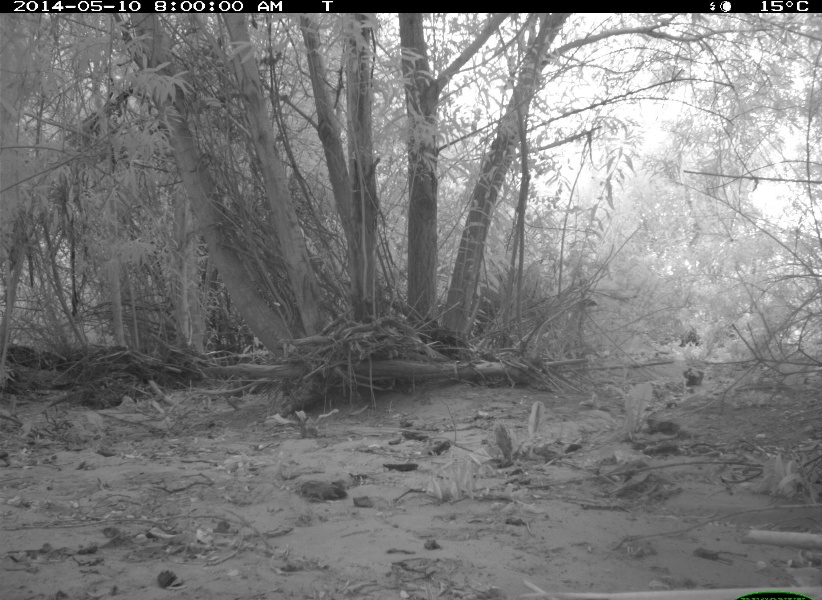}}
\hspace{1mm}
\subfigure[EffNet-B0: 0.42, EffNet-B3: 0.57, MNetV2-224: 0.12, MNetV2-320: 
0.80, SSDLite: 0.16, EffDet-D0: 0.10]{\includegraphics[width=0.32\textwidth]{
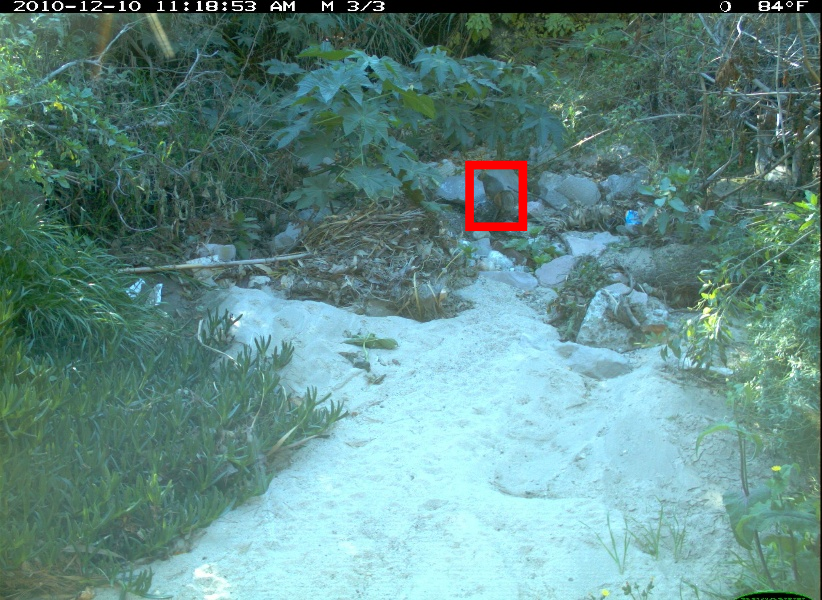}}
\hspace{1mm}
\subfigure[EffNet-B0: 0.99, EffNet-B3: 0.99, MNetV2-224: 0.99, MNetV2-320: 
0.99, SSDLite: 0.87, EffDet-D0: 0.88]{\includegraphics[width=0.32\textwidth]{
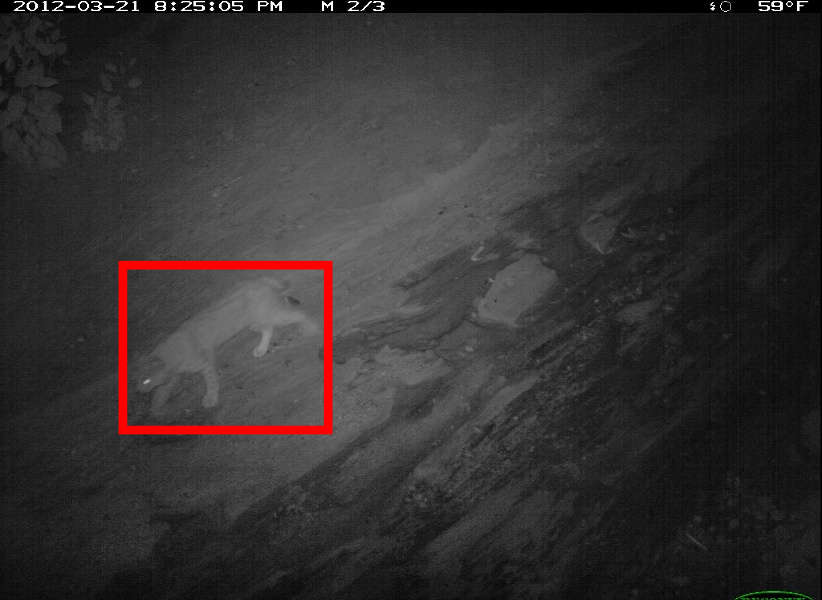}}
\subfigure[EffNet-B0: 0.13, EffNet-B3: 0.03, MNetV2-224: 0.29, MNetV2-320: 
0.31, SSDLite: 0.15, EffDet-D0: 
0.05]{\includegraphics[width=0.32\textwidth]{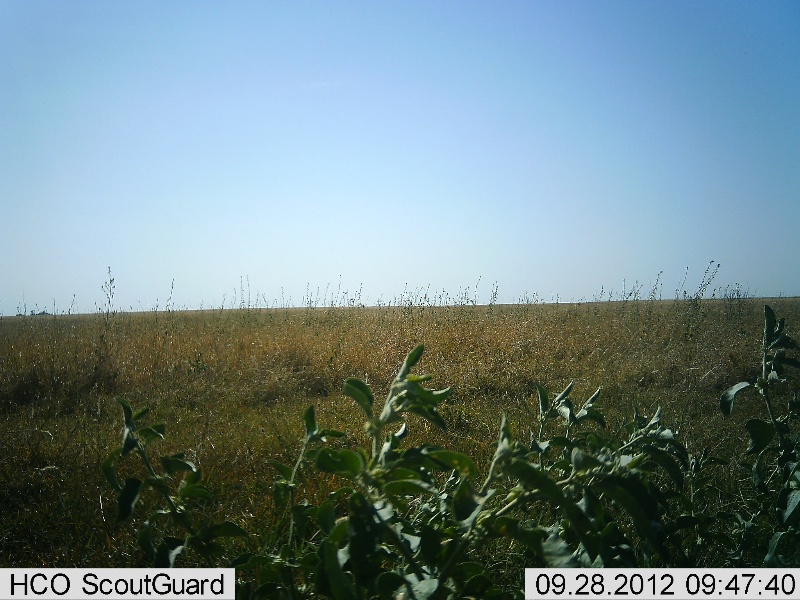}}
\hspace{1mm}
\subfigure[EffNet-B0: 0.99, EffNet-B3: 0.99, MNetV2-224: 0.99, MNetV2-320: 
0.99, SSDLite: 0.90, EffDet-D0: 
0.94]{\includegraphics[width=0.32\textwidth]{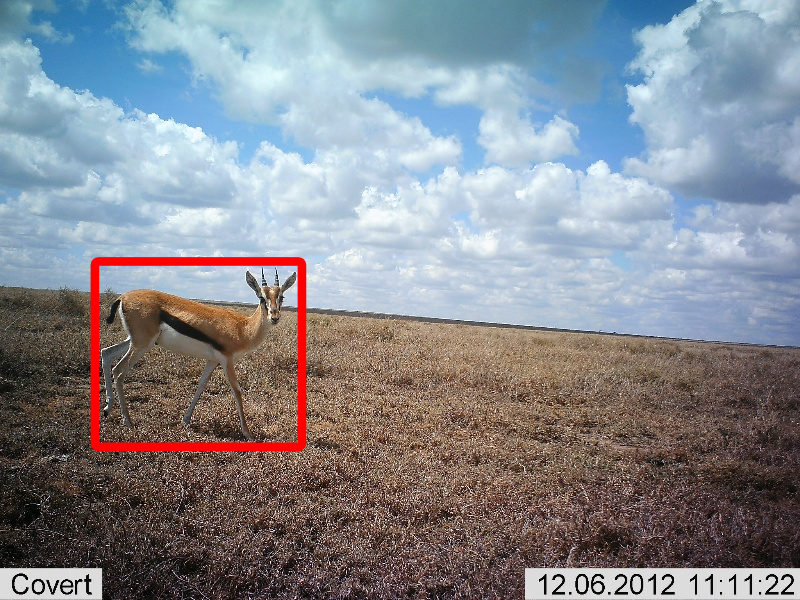}
}
\hspace{1mm}
\subfigure[EffNet-B0: 0.22, EffNet-B3: 0.66, MNetV2-224: 0.48, MNetV2-320: 
0.91, SSDLite: 0.77, EffDet-D0: 
0.70]{\includegraphics[width=0.32\textwidth]{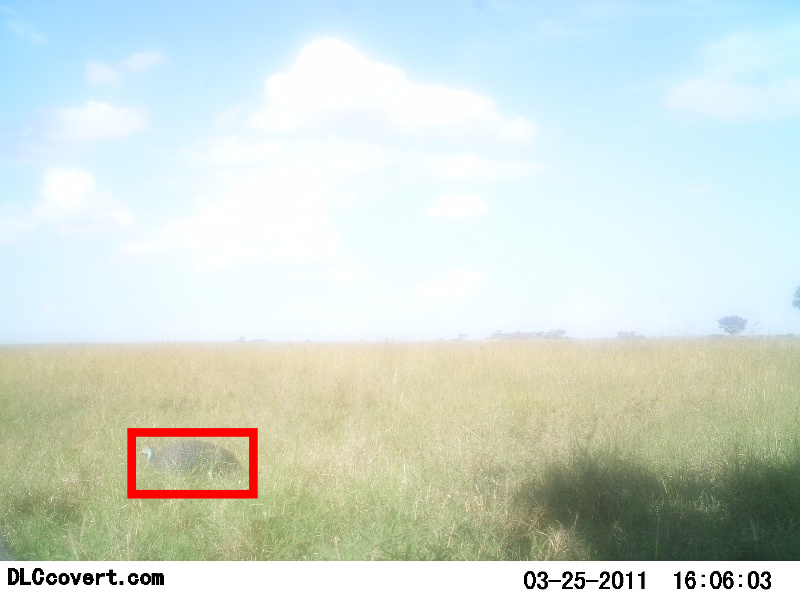}
}
\end{center}
   \caption{Sample classification results reached by the investigated models
highlighting their confidence for the nonempty class. The top row depicts images 
from the Caltech dataset and the bottom row from the Snapshot Serengeti dataset. 
The first column depicts instances of the empty class while the remainder are 
nonempty images with the animals highlighted.}
\label{fig:predictions}
\end{figure*}

We used a Raspberry Pi 3 model B running Raspbian GNU/Linux 10 as a reference 
edge device to assess the models' latency. The models were converted to the 
TensorFlow Lite format without quantization and their latency was calculated 
using the native benchmark tool provided by 
TensorFlow\footnote{\url{https://www.tensorflow.org/lite/performance/measurement}}. The reported latency values were calculated as the average over 50 runs for
each model. Additionally, we also evaluate the models version obtained as a 
result of post-training quantization~\cite{jacob2018quantization}. In this 
scenario, a subset of 500 training instances was employed to calibrate the 
operations to work with the integer type, while maintaining model inputs and 
outputs as floating point to keep the original model interface.

\section{Experimental Results}

\subsection{Classifiers vs. Detectors}

In order to compare the performance of classifiers and detectors as fairly as 
possible, the models were trained using subsets composed by instances of the
nonempty class annotated with bounding boxes. Despite of the fact that these 
subsets were generated using a much smaller amount of the total images available 
for classification, leading to possible sub-optimal models, this number of 
instances can provide a more realistic perspective of the problem. Indeed, as 
pointed out by Schneider \etal~\cite{schneider2020three}, the vast majority of 
small-scale research projects focused on camera trap do not have a large amount 
of labeled images.

\textbf{Results:} Figure~\ref{fig:pr_curve} shows the precision-recall curves of
the investigated models for each dataset. As expected, detectors outperformed 
classifiers in all datasets, especially Caltech. Considering a confidence 
threshold producing 96\% of recall, EfficientDet-D0 was able to eliminate more 
than twice as many empty images when compared to the classifiers using Caltech 
dataset. In terms of SSDLite+MobileNetV2, it also obtained a significantly 
higher true negative rate (at least 19\% in absolute values), as reported in 
Table~\ref{table:precision_tvn}. The scenario is quite similar for the Snapshot
Serengeti dataset, since detectors also outperformed classifiers by a 
significant margin (at least 8\% of precision). Figure~\ref{fig:predictions}
shows some images to illustrate the results attained in our experiments.

\begin{figure*}
\begin{center}
\subfigure[SS-Site]{\includegraphics[width=0.42\textwidth]{
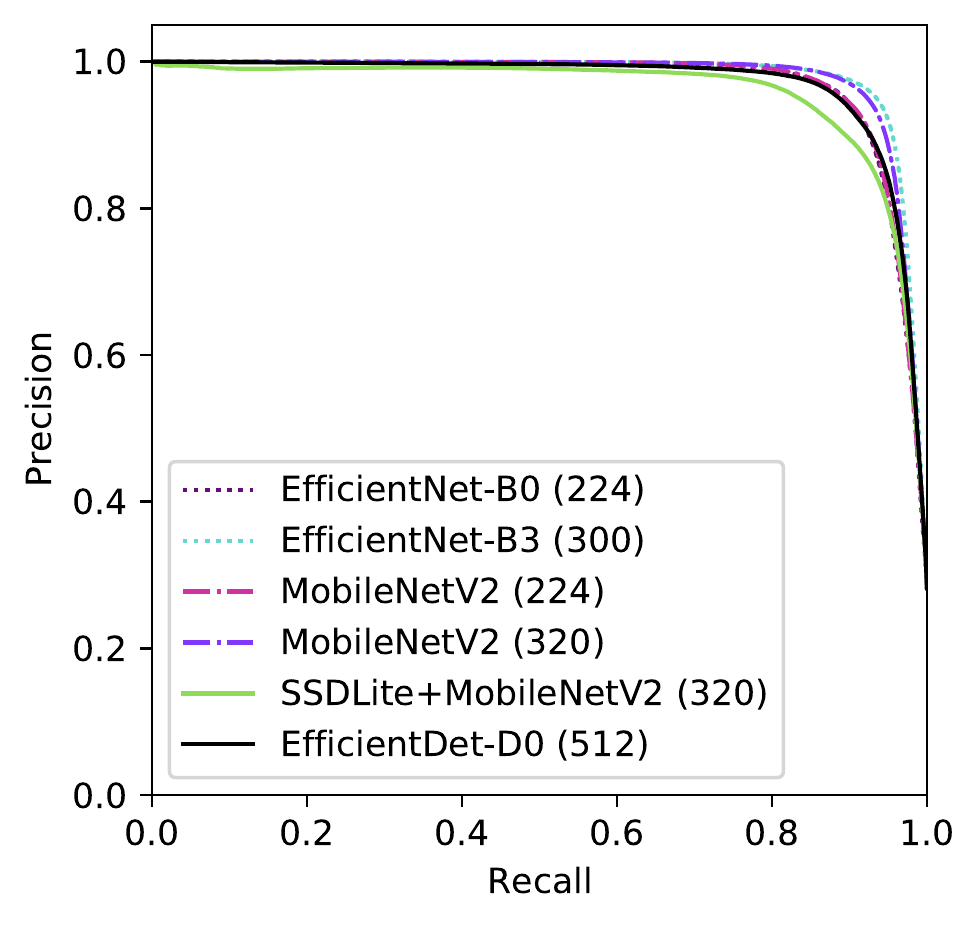}}
\hspace{0.05\textwidth}
\subfigure[SS-Time]{\includegraphics[width=0.42\textwidth]{
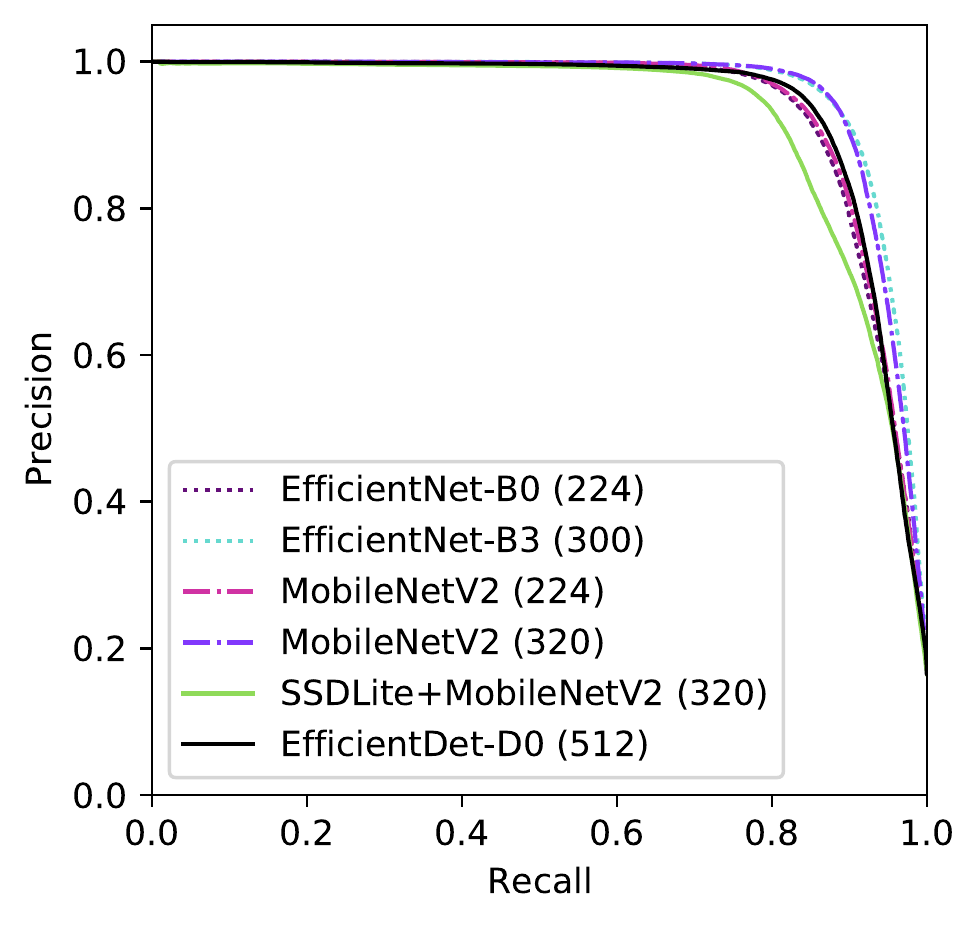}}
\end{center}
   \caption{Precision-recall curve for the nonempty class of classifiers trained 
in sets with ten times more images from the SS dataset. The curves of the
detectors refer to the models trained on the original smaller set. Best viewed 
in color.}
\label{fig:pr_curve_more_images}
\end{figure*}

The poor classifier performance on the Caltech dataset can be due to several
factors, such as the low variability of backgrounds, the size of animals, 
camouflage, quality of images, among others. However, our experiments were not 
designed to identify these nuances intrinsic to this dataset. On the other hand, 
this may be an interesting topic to be investigated in future work.

\textbf{Latency-precision trade-off:} Table~\ref{table:precision_tvn} shows 
the latency for the inference of each model. Although EfficientDet-D0 offers a 
strong baseline for the problem, its latency is more than four seconds, which is
prohibitive because camera trap images are usually obtained within one second 
between them. In this context, SSDLite+MobileNetV2 may be deemed to attain
superior performance since its latency is below one second and it eliminated 
at least 10\% more images than the classifiers with comparable latencies.

\subsection{Training with more Images}

A common way to improve model performance is to use more training 
instances~\cite{bit2020Kolesnikov}. Following this strategy, an experiment was
carried out to train the classifiers using the subsets of the SS dataset. This 
dataset contains ten times more images than the one used in the previous 
experiment. Figure~\ref{fig:pr_curve_more_images} and 
Table~\ref{table:more_images} summarize the results attained in this scenario. 
The results indicate that classifiers outperformed detectors. We can highlight 
that MobileNetV2 (224) reached performance similar to SSDLite+MobileNetV2, with 
less than half of inference latency though. It is important to note, however, 
that all detectors were trained using data subsets from the previous experiment. 
Therefore, in order to provide a fair comparison, a similar amount of training 
instances used to train the classifiers should be used to train the detectors. 
On the other hand, obtaining more annotated instances for the detection problem 
is expensive. This is the reason we do not show results using the same dataset 
for training classifiers and detectors. However, we can conclude based on the 
results obtained in this experiment that, depending on the number of labeled 
instances available, classifiers can be a viable option for the problem of 
identifying empty images, also showing better efficiency in terms of 
computational resources. 

\begin{table}[htb]
\begin{center}
\setlength{\tabcolsep}{3pt}
\begin{tabular}{l|cc|cc}
 \hline
\textbf{}                            & \multicolumn{2}{c|}{\textbf{SS-Site}}    
 
                                & \multicolumn{2}{c}{\textbf{SS-Time}}          
                          \\
\multicolumn{1}{c|}{\textbf{Model}} & \multicolumn{1}{c}{\textbf{Precision}} & 
\multicolumn{1}{c|}{\textbf{TNR}} & \multicolumn{1}{c}{\textbf{Precision}} & 
\multicolumn{1}{c}{\textbf{TNR}} \\ \hline \hline
Efficientnet-B0                      & 73.92\%                               & 
86.78\%                           & 48.81\%                               & 
80.25\%                          \\
Efficientnet-B3                      & \textbf{87.67\%}                      & 
\textbf{94.73\%}                  & \textbf{64.28\%}                      & 
\textbf{89.54\%}                 \\
MNetV2-224                      & 75.18\%                               & 
87.63\%                           & 49.12\%                               & 
80.50\%                          \\
MNetV2-320                      & \textbf{82.89\%}                      & 
\textbf{92.26\%}                  & \textbf{58.61\%}                      & 
\textbf{86.70\%}                 \\ \hline
\end{tabular}
\end{center}
\caption{Comparison of precision for the nonempty class and the true negative 
rate (TNR), with a recall at 96\%, for classifiers trained in sets with ten 
times more images from the SS dataset. Note the significant improvement in the 
performance of the classifiers, exceeding by a large margin the detectors 
trained in the original set.}
\label{table:more_images}
\end{table}

\begin{figure*}
\begin{center}
\subfigure[SS-Site]{\includegraphics[width=0.49\textwidth]{
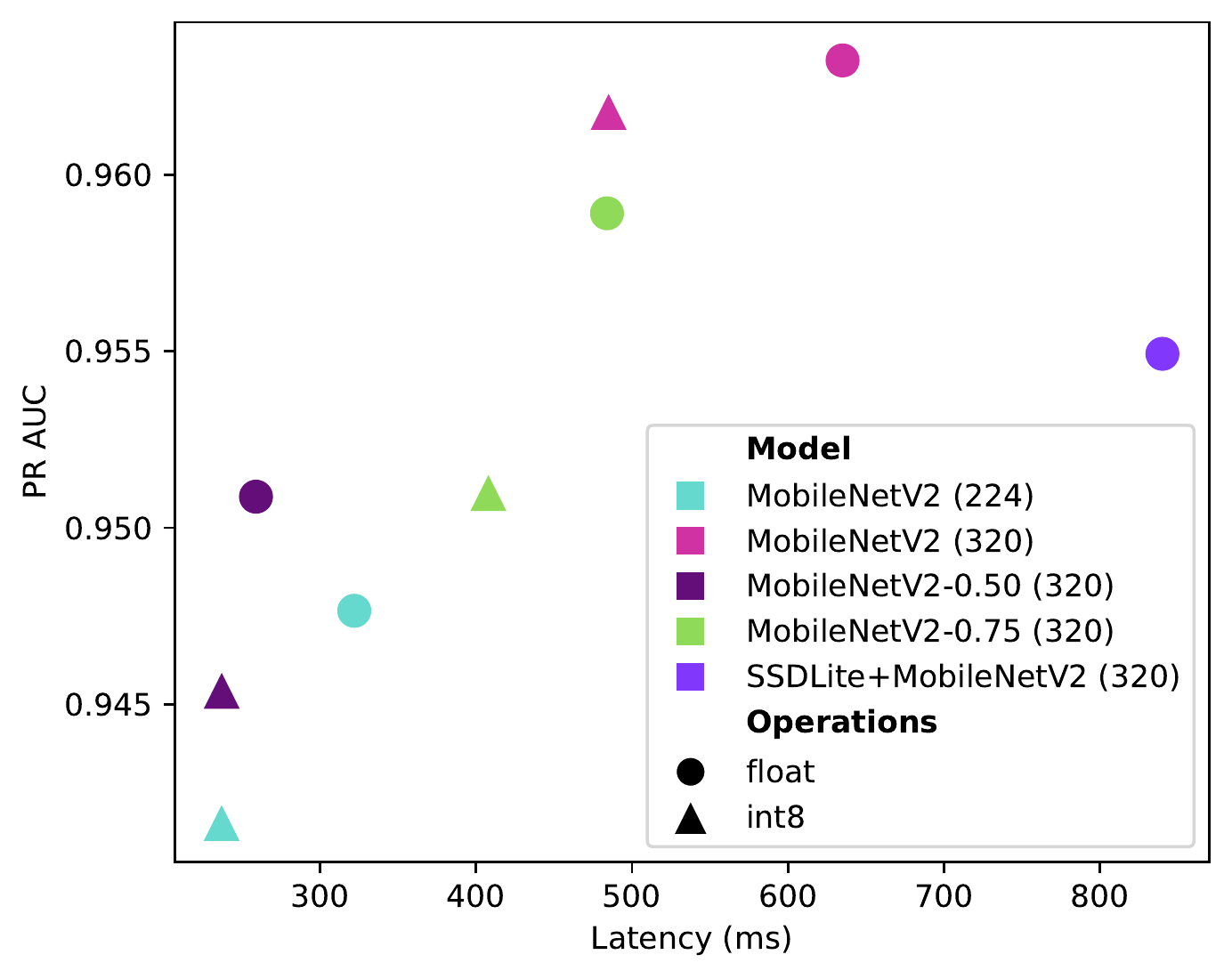}}
\hspace{0.01\textwidth}
\subfigure[SS-Time]{\includegraphics[width=0.49\textwidth]{
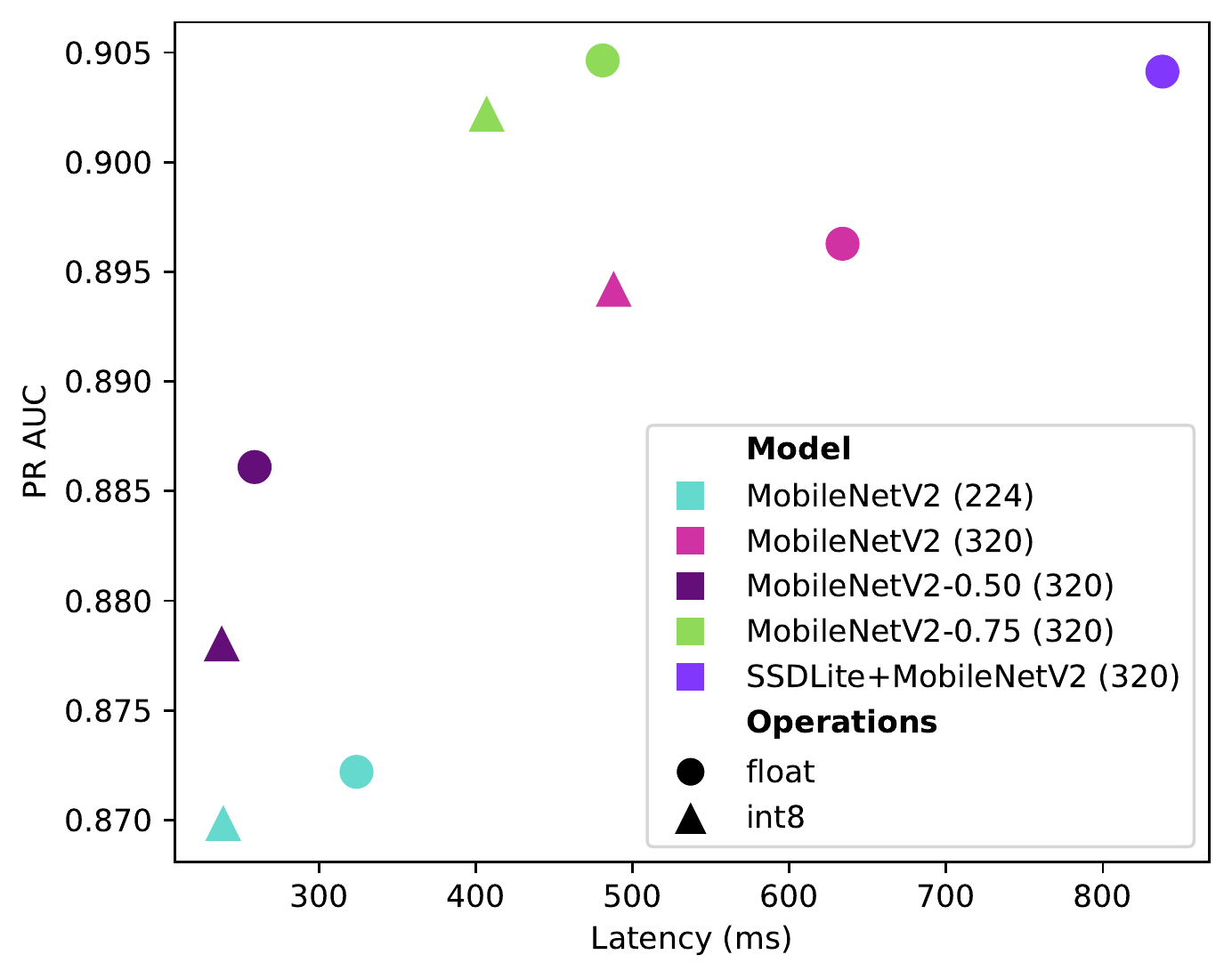}}
\end{center}
   \caption{Latency vs. area under the precision-recall curve (PR AUC) for
various versions of MobileNetV2 in the SS. The results refer to the models 
converted to TensorFlow Lite (floating point and integer) and evaluated in a 
subset of 5,000 test instances. Best viewed in color.}
\label{fig:latency_prauc_ss}
\end{figure*}

\subsection{Model Optimization}

\begin{table*}[htb]
\begin{center}
\begin{tabular}{ll|c|c|cc|cc}
\hline
\textbf{}       &      &             &                & 
\multicolumn{2}{c|}{\textbf{SS-Site}}                                  & 
\multicolumn{2}{c}{\textbf{SS-Time}}                                 \\
\multicolumn{2}{c|}{\textbf{Model}} & \textbf{CPU} & \textbf{Memory}  & 
\multicolumn{1}{c}{\textbf{Precision}} & \multicolumn{1}{c|}{\textbf{TNR}} & 
\multicolumn{1}{c}{\textbf{Precision}} & \multicolumn{1}{c}{\textbf{TNR}} \\
\hline \hline
\multirow{3}{*}{MobileNetV2-224}  & Float       & 322ms  &  22.7MB    & 58.20\%

         & 73.07\%                           & 25.67\%                          
 
 & 45.33\%                          \\
                      & Int8        & \textbf{237ms} & \textbf{9.3MB} & 55.52\%

         & 69.99\%                           & 26.18\%                          
 
 & 46.74\%                          \\
                      & Int8 (quant aware)        & 239ms & 9.8MB 
 & 57.31\% & 72.10\%  & - & -   \\ \hline
\multirow{2}{*}{MobileNetV2-0.50-320} & Float       & 259ms  & 19.7MB     & 
54.86\%

         & 69.15\%                           & 27.76\%                          
 
 & 50.86\%                          \\
                      & Int8        & 237ms  &  9.6MB        & 51.50\%

         & 64.70\%                           & 26.67\%                          
 
 & 48.09\%                          \\ \hline
\multirow{3}{*}{MobileNetV2-0.75-320} & Float       & 484ms & 19.8MB   & 
64.89\%

         & 79.72\%                           & \textbf{31.37\%}                 
 
 & \textbf{58.71\%}                 \\
                      & Int8        & 408ms   & 13.3MB       & 59.50\%

         & 74.49\%                           & 30.44\%                          
 
 & 56.87\%                          \\
                      & Int8 (quant aware)        & 413ms & 13.4MB 
 & 61.38\% & 76.41\%  & - & -   \\ \hline
\multirow{2}{*}{MobileNetV2-320} & Float       & 635ms  & 33.5MB    & 66.65\%

         & 81.25\%                           & 30.34\%                          
 
 & 56.65\%                          \\
                      & Int8        & 485ms & 13.9MB  & \textbf{66.70\%}   
         & \textbf{81.28\%}                  & 30.98\%                          
 
 & 57.95\%                          \\ \hline
\multirow{2}{*}{SSDLite+MobileNetV2} & Float       & 840ms & 31.7MB   & 73.68\%

         & 86.62\%                           & 39.50\%                          
 
 & 71.09\%                          \\
                      & Int8        & 575ms  &  13.6MB       & 36.83\%         
 
         & 35.72\%                           & 19.27\%                          
 
 & 20.90\%                          \\ \hline
\end{tabular}
\end{center}
\caption{Performance comparison of MobileNetV2 models for various widths on 
the SS dataset. The models were converted to TensorFlow Lite and evaluated using 
a sample of 5,000 instances randomly chosen from the validation set.}
\label{table:resolution}
\end{table*}

It is possible to observe from results shown in previous sections that models
dealing with higher image input resolutions achieved higher performance but at 
the expense of greater inference latency. In order to reduce latency of these 
models, we performed an experiment using the SS dataset and MobileNetV2 (320). 
In this experiment, the number of filters was reduced by adjusting the alpha 
parameter. We also evaluated the post-training quantization of the models. Since 
the swish activation function used by EfficientNets is not very well supported 
for quantization~\cite{xiong2020mobiledets}, models of this family were not 
used here. In addition, models evaluation was carried out on a sample of 5,000 
instances obtained from the validation set, because the TensorFlow Lite
quantized models are not optimized for inference on x86 architectures and 
graphics cards used for training.

The results shown in Figure~\ref{fig:latency_prauc_ss} reinforce the role of 
input resolution, since a MobileNetV2 with half the filters (MobileNetV2-0.50) 
but higher resolution (320) obtained overall performance superior to the
performance of the standard model (MobileNetV2 (224)). In 
Table~\ref{table:resolution}, we may observe that it is possible to reduce about 
23\% of model latency due to both quantization (SS-site) and number of filters 
reduction (SS-time), maintaining compatible performance. It is worth mentioning 
that the quantized version of the SSDLite + MobileNetV2 detector did not obtain 
good results at the 96\% recall level used. These results indicate the 
importance of assessing the performance of the models after quantization, 
especially when the confidence threshold is adjusted at certain levels. One way 
to avoid this problem would be quantization-aware training.

To assess this, we fine tuned the models MobileNetV2-224 and 
MobileNetV2-0.75-320 trained on SS-Site using quantization-aware training 
for 2 epochs with the initial learning rate divided by 10. In this case, we were
able to improve the results of the quantized versions, as shown in 
Table~\ref{table:resolution}. Unfortunately, we were not able to assess SSDLite 
in the same way due to the fact that Tensorflow Object Detection API did not 
support quantization-aware training for TF2 at the experiments' time.

Although the focus of this work is the latency-precision trade-off, we also 
measured the memory usage of each model, as shown in 
Table~\ref{table:resolution}. Quantized models have a significant reduction in 
memory usage compared to the floating point models. However, when observed the
amount of RAM available on Raspberry Pi (1GB), the memory required by the 
most complex model is very low (33.5MB), therefore, not affecting the
performance of other processes that may be running on the device.

\section{Conclusion}

In this work, we presented a comparative study between detection and
classification models in the context of identification of nonempty images of 
animals on edge devices. Our results showed that detection models outperform 
classifiers when both are trained using the same training set, but their 
superior inference latency can limit their use on edge devices. Moreover,  
depending on the dataset, it is possible to train classifiers to obtain 
satisfactory results, especially when there is massive number of images 
available, as in the Snapshot Serengeti dataset. When detector is essential, 
e.g. Caltech dataset, but the model's latency is not within the design 
requirements, it may be necessary to use hardware accelerators, such as EdgeTPUs 
or DSPs. Another limitation to detectors is the difficulty on obtaining new 
instances annotated with bounding boxes, which is time-consuming and expensive. 
In this situation, one possibility is to use techniques tackling few or no 
labels, such as semi-supervised learning and self-supervised learning. Other 
alternative is training an agnostic detection model designed to run on edge 
devices, inspired by MegaDetector. Regarding the optimization strategies 
evaluated, the post-training quantization proved to be effective, but it 
requires a careful evaluation of the resulting model, which may decrease its 
performance due to the quantization process. However, quantization-aware 
training may solve this issue. Finally, using models with fewer filters but with 
higher resolution was also effective. Therefore, techniques such as knowledge 
distillation and model pruning are interesting directions for future work.

\vspace{8mm}

\noindent\textbf{Acknowledgements:} This research, according to 
Article 48 of Decree nº 6.008/2006, was partially funded by Samsung Electronics 
of Amazonia Ltda, under the terms of Federal Law nº 8.387/1991, through 
agreement nº 003/2019, signed with ICOMP/UFAM. This study was supported by the 
Foundation for Research Support of the State of Amazonas (FAPEAM) - POSGRAD 
Project, and the Coordination for the Improvement of Higher Education Personnel 
- Brazil (CAPES) - Finance Code 001. The funders had no role in study design, 
data collection and analysis, decision to publish, or preparation of the 
manuscript.

{\small
\bibliographystyle{ieee_fullname}
\bibliography{egbib}
}

\end{document}